\documentclass[letterpaper]{article} % DO NOT CHANGE THIS
\usepackage{aaai25}  % DO NOT CHANGE THIS
\usepackage{times}  % DO NOT CHANGE THIS
\usepackage{helvet}  % DO NOT CHANGE THIS
\usepackage{courier}  % DO NOT CHANGE THIS
\usepackage[hyphens]{url}  % DO NOT CHANGE THIS
\usepackage{graphicx} % DO NOT CHANGE THIS
\usepackage{amsmath, amssymb}
\usepackage{booktabs} % 确保横线正常显示
\usepackage{natbib}  % DO NOT CHANGE THIS AND DO NOT ADD ANY OPTIONS TO IT
\usepackage{caption} % DO NOT CHANGE THIS AND DO NOT ADD ANY OPTIONS TO IT
\usepackage{algorithm}
\usepackage{algpseudocode}
\usepackage{subcaption} 
\usepackage{newfloat}
\usepackage{listings}
\DeclareCaptionStyle{ruled}{labelfont=normalfont,labelsep=colon,strut=off} % DO NOT CHANGE THIS
\floatstyle{ruled}
\newfloat{listing}{tb}{lst}{}
\floatname{listing}{Listing}
\title{Leveraging Multi-modal Representations to Predict Protein Melting Temperatures}
\author{
    Daiheng Zhang\textsuperscript{\rm 1},
    Yan Zeng\textsuperscript{\rm 1},
    Xinyu Hong\textsuperscript{\rm 1},
    Jinbo Xu\textsuperscript{\rm 1}
}
\affiliations{
    \textsuperscript{\rm 1}MoleculeMind Ltd., Beijing, China\\
    dz367@rutgers.edu, zy1104865009@gmail.com, xinyuhong@moleculemind.com, jinboxu@gmail.com
}

\usepackage{bibentry}
\begin{document}

\maketitle

\begin{abstract}
Accurately predicting protein melting temperatures ($\Delta T_m$) is fundamental for assessing protein stability and guiding protein engineering. Leveraging multimodal protein representations has shown great promise in capturing the complex relationships among protein sequences, structures, and functions. In this study, we develop models based on powerful protein language models—including ESM2, ESM3, SaProt, and AlphaFold—using various feature extraction methods to enhance prediction accuracy. By utilizing the ESM3 model, we achieve a new state-of-the-art performance on the s571 test dataset, obtaining a Pearson correlation coefficient (PCC) of 0.50. Furthermore, we conduct a fair evaluation to compare the performance of different protein language models in the $\Delta T_m$ prediction task. Our results demonstrate the strength of integrating multimodal protein representations could advance the prediction of protein melting temperatures. 
\end{abstract}


\section{Introduction}
Proteins play a pivotal role in various biological applications, such as catalyzing biochemical reactions, immune function, and metabolism regulation. Composed of sequences built from 20 different classes of amino acids, proteins fold into complex structures—both sequence and structure determine their functions \citep{whisstock2003prediction}. Therefore, exploring appropriate protein representations is crucial for related tasks. Large-scale protein language models (PLMs) have demonstrated excellent performance in protein representation capabilities \citep{bepler2021learning,lin2022language,hayes2024simulating,su2023saprot}. The pre-training strategies enhance the models' ability to capture nuanced features and patterns in protein sequences and structures, effectively transferring to downstream tasks like understanding protein fitness \citep{ouyang2024predicting,chen2024end} and evolutionary dynamics \citep{hie2024efficient}.

With stabilized structures, downstream engineering of proteins becomes more feasible. Mutations are commonly used in protein engineering to study and improve protein structure and function, making the accurate quantification of mutation effects crucial for studying the evolutionary fitness of proteins \citep{pandurangan2020prediction}. Thermodynamic stability \citep{pires2014mcsm,umerenkov2022prostata,benevenuta2021antisymmetric,pandurangan2020prediction} and enzyme kinetic parameters \citep{li2022deep,yu2023unikp} are widely explored in mutation-related tasks. Benefiting from deep mutational scanning (DMS) databases containing protein fitness data \citep{fowler2014deep,tsuboyama2023mega}, thermodynamic stability ($\Delta \Delta G$) prediction has been extensively studied, and its performance has greatly improved. However, the prediction of changes in melting temperature ($\Delta T_m$) has been less explored compared to $\Delta \Delta G$ prediction. Deep learning-based methods are largely absent in addressing this problem \citep{xu2023improving,masso2014auto,masso2008accurate,pucci2016predicting}, partly due to a lack of experimental data and partly because the issue has not received significant attention.

In this paper, we propose a new prediction framework, \textit{ESM3-DTm}, for $\Delta T_m$. We fine-tune three distinct protein language models—ESM2 \citep{lin2022language}, ESM3 \citep{hayes2024simulating}, and SaProt \citep{su2023saprot}—and also explore using OpenFold \citep{ahdritz2024openfold} to extract features by incorporating different regression heads into the architecture. Among these approaches, we found that using \textit{ESM3-DTm} accepting both sequence and structure to obtain embeddings yielded the best results, achieving state-of-the-art (SOTA) performance with a Pearson correlation coefficient (PCC) of 0.50, mean absolute error (MAE) of 5.21, and root mean square error (RMSE) of 7.68. We also demonstrate the impact of different finetuning methods on the results.

\section{Preliminary}
\subsection{Problem Setup} A protein $P = (a_1, a_2, \dots, a_L)$ is a sequence of amino acids, where each $a_i \in AA$, and $AA = {A, C, \dots, Y}$ represents the 20 standard amino acid types. Let $\mu = (w, m)$ denote a mutation that substitutes the amino acid at position $w$ in $P$ with amino acid type $m \in AA$. Our goal is to predict the change in melting temperature $\Delta T_m \in \mathbb{R}$ for the protein $P$ resulting from the mutation $\mu$.

\subsection{Protein large language models}
Over recent years, large language models have played an ever more significant role in protein research, providing innovative insights and enhanced abilities for comprehending and modifying proteins\cite{zhang2024scientific}. Most of them are encoder-only models, which are built upon the encoder of Transformer, enables the encoding of protein
sequences or structures into fixed-length vector representations. From this series of mainstream models, we selected ESM2 and SaProt to further explore their representation capabilities. ESM2 is one of the largest architectures among single sequence models and stands out for its role in structure prediction. We adopt the architecture with 640 million parameters and 36 layers. SaProt is a bilingual protein language model featuring structure-aware embeddings, undergoes training on Foldseek’s 3Di structures\cite{van2022foldseek} and amino acid sequences. We used the same settings for SaProt as we did for ESM2. In addition to encoder-only models, we also explored encoder-decoder models. The advantage of having a decoder is that it provides the model with strong generative capabilities. Here, we investigated ESM3, a newly released co-design multimodal model. We used the publicly available version with 1.4 billion parameters. Apart from these language models, AlphaFold\cite{jumper2021highly} has shown its highly effective in predicting protein structures from sequences by leveraging evolutionary information through multiple sequence alignments (MSA). It can also be used for feature extraction. Therefore, we also utilized OpenFold as a backbone for further experiments.

\subsection{Data} To compare our method with existing models, we use the training and test datasets proposed by \textbf{GeoStab}. The training set, s4346, comprises 4,346 single-point mutations across 349 proteins, collected from ProThermDB \cite{gromiha1999protherm, gromiha2000protherm, gromiha2002protherm, kumar2006protherm} and ThermoMutDB \cite{xavier2021thermomutdb}, both of which are dedicated $\Delta T_m$ databases. The test set, s571, consists of 571 single-point mutations across 37 proteins, also collected from the same sources.

We observed that the baseline method lacks a train/validation split within the training set, which can easily lead to overfitting. To address this issue, we used MMseqs2\cite{steinegger2017mmseqs2} at 50\% sequence identity and then split it into training and validation sets in an 8:2 ratio. Hyperparameter tuning was conducted using this split. After identifying the best hyperparameters, we retrained the model on the combined training and validation sets, aligning with the train-test split setting of the previous baseline \textbf{GeoStab}.

The original dataset contains only sequence data. For input to the OpenFold backbone, multiple sequence alignments (MSAs) are required; we computed these using ColabFold \cite{mirdita2022colabfold}. For the ESM3 backbone with PDB input option and for the SaProt backbone input, PDB structures are needed. Our PDB structure dataset consists of two parts: for proteins with available PDB IDs, we retrieved the corresponding structures from the Protein Data Bank (PDB); for proteins with only UniProt IDs and for all mutated structures, we generated PDB structures using ColabFold.

% \begin{figure*}[!t]
%     \centering
%     % First image
%     \label{fig:openfold_esm}
%     \includegraphics[width=1.0\textwidth]{AnonymousSubmission/LaTeX/openfold_esm.pdf}
    
%     \vspace{0.5em} % Optional: Adjusts the space between the two images
    
%     % Second image
%     \label{fig:ESM3}
%     \includegraphics[width=0.8\textwidth]{AnonymousSubmission/LaTeX/esm3.pdf}
    
%     \caption{
%         \textbf{Model architecture}. \textit{ESM3-DTm} efficiently predicts $\dTm$. We also present \textit{ESM2-DTm, Saprot-DTm} and \textit{Openfold-DTm} here. ''I4A'' means mutation from I to A at position 4.
%     }
    
% \end{figure*}
\begin{figure*}[!t]
    \centering
    % First sub-figure
    \begin{subfigure}[b]{0.9\textwidth}

        \centering
        \includegraphics[width=\textwidth]{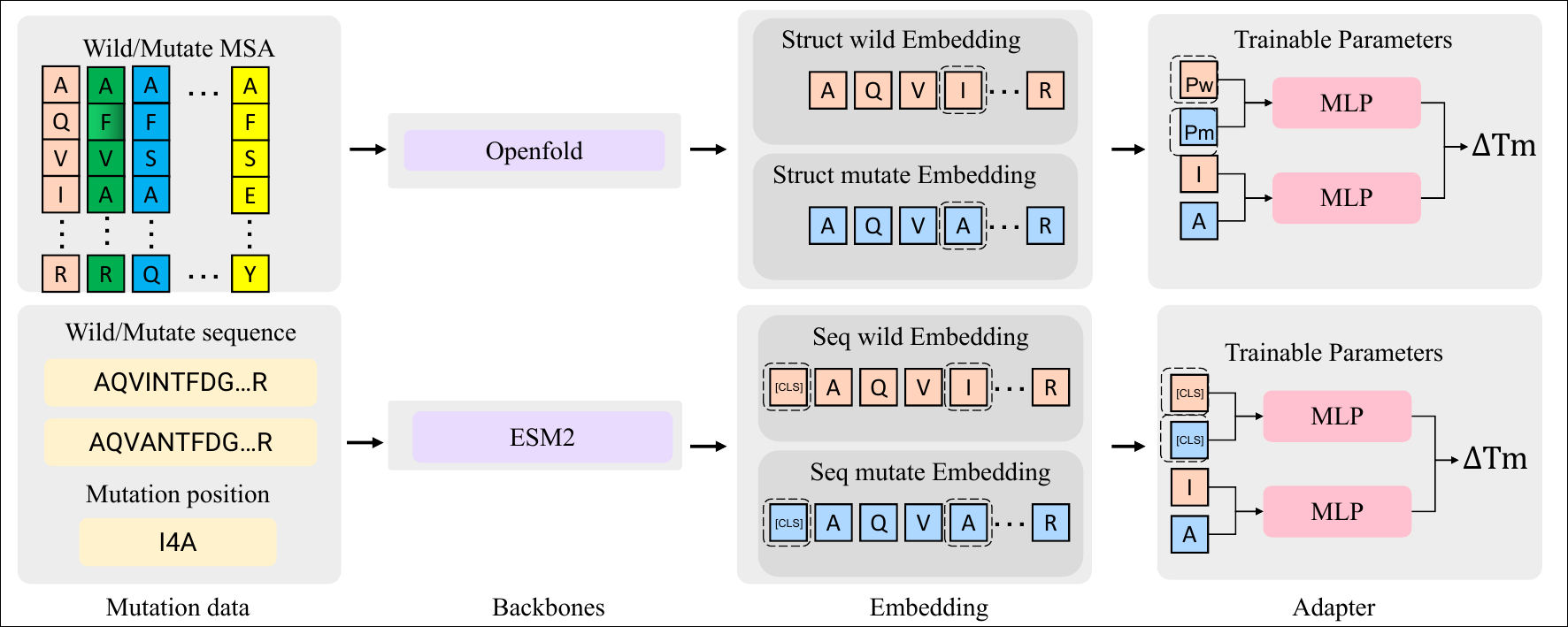}
        \caption{Openfold and ESM2 Backbone Architecture}
        \label{fig:openfold_esm}
    \end{subfigure}
    
    \vspace{0.5em} % Optional: Adjusts the space between the two sub-figures
    
    % Second sub-figure
    \begin{subfigure}[b]{0.9\textwidth}

        \centering
        \includegraphics[width=\textwidth]{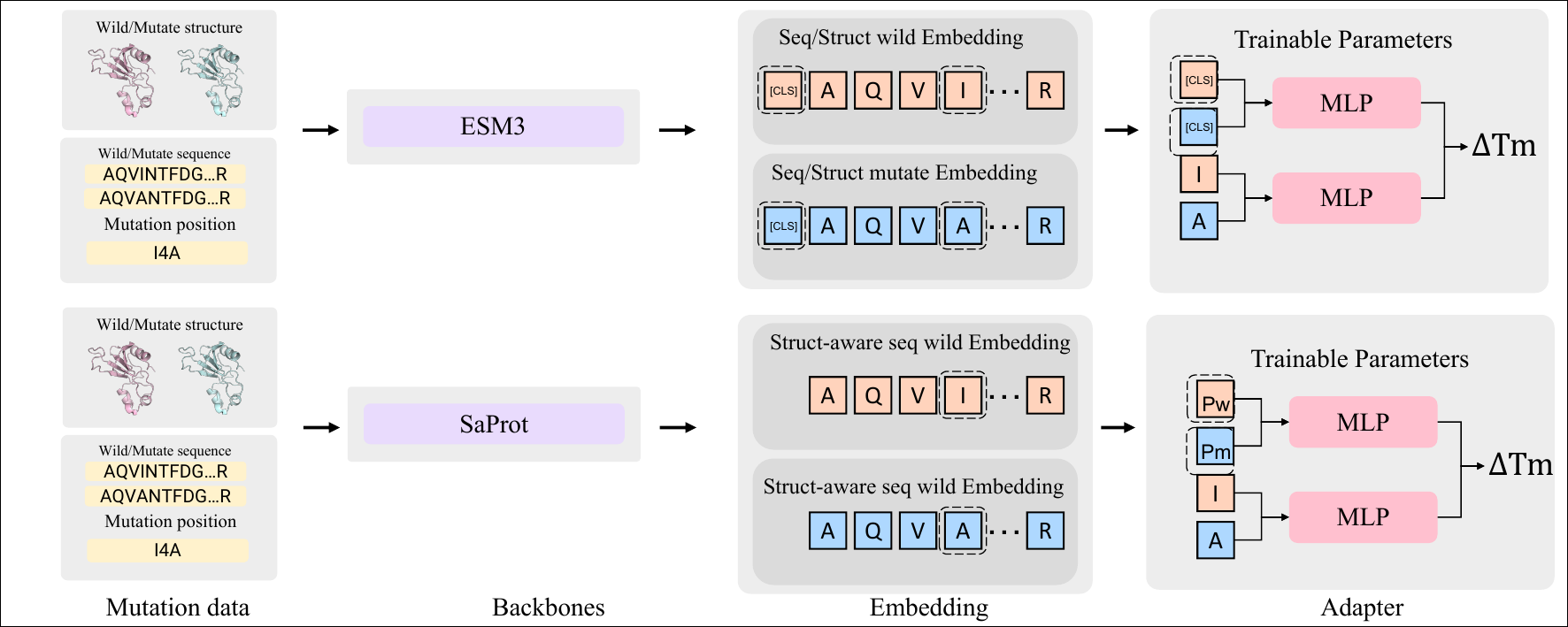}
        \caption{ESM3 and SaProt Backbone Architecture}
        \label{fig:ESM3}
    \end{subfigure}
    
    \caption{
        \textbf{Model Architecture}. \textit{ESM3-DTm} efficiently predicts $\Delta T_m$. We also present \textit{ESM2-DTm}, \textit{Saprot-DTm}, and \textit{Openfold-DTm} here. ``I4A'' means mutation from I to A at position 4.
    }
    \label{fig:model_architecture}
\end{figure*}

\section{Experiments}
\subsection{Model Setup and Implementation Details}

For each mutation $\mu = (w, m)$, where the amino acid at position $w$ is substituted with amino acid type $m \in AA$, we denote $P_w$ and $P_m$ as the representations of the entire wild-type and mutated protein sequences, respectively. We use $a_w$ and $a_m$ to represent the embeddings at the specific mutated position in the wild-type and mutated proteins. We denote $cls_w$ and $cls_m$ as the CLS token representations of the wild-type and mutated proteins.

We treat the prediction of the mutation effect on $\Delta T_m$ as a regression task involving two sequences: the wild-type and the mutated protein. Our model is built upon a protein language model backbone that accepts inputs of both wild-type and mutated protein sequences, along with structure-related information. In our approach, protein language models serve as feature extractors.

Our most powerful model, \textit{ESM3-DTm}, is built upon ESM3-1.4B \cite{hayes2024simulating}, a multimodal protein language model that accepts both sequence and PDB structure inputs, as illustrated in Figure \ref{fig:ESM3}. We extract sequence and structure features by applying a linear layer to the final hidden layer outputs from ESM3. We then concatenate the embeddings from $Struct_{\text{cls\_w}}$ and $Seq_{\text{cls\_w}}$ to obtain $cls\_w$, and similarly concatenate $Struct_{\text{cls\_m}}$ and $Seq_{\text{cls\_m}}$ to obtain $cls_m$. We obtain $a_w$ and $a_m$ in the same manner. Finally, we feed these into a regression head and average the predictions from the individual models in the ensemble. We design various regression heads in Section \ref{Re_head} to predict $\Delta T_m$. The detailed progress is explained in algorithm \ref{tab:ESM_dtm}.

\begin{algorithm}[!t]
\caption{\textit{ESM3-DTm} Model}
\label{tab:ESM_dtm}
\begin{algorithmic}[1]  % ✅ 启用 algorithmic 环境

\State \textbf{Input:} CLS token embedding for wild-type $CLS_{\mathrm{w}} \in \mathbb{R}^d$,
mutated $CLS_{\mathrm{m}} \in \mathbb{R}^d$; mutated position token embedding for wild-type $a_{\mathrm{w}} \in \mathbb{R}^d$, mutated $a_{\mathrm{m}} \in \mathbb{R}^d$

\State \textbf{Step 1: Regression Heads}
\State Head1 = Flatten($\mathbf{a}_{\mathrm{m}} \otimes \mathbf{a}_{\mathrm{w}} \in \mathbb{R}^{d^2}$) $\xrightarrow{\mathbf{W}} \mathbb{R}^{d}$
\State Head2 = LayerNorm($\mathbf{CLS}_{\mathrm{w}} - \mathbf{CLS}_{\mathrm{m}}$) $\oplus$ LayerNorm($\mathbf{a}_{\mathrm{w}} - \mathbf{a}_{\mathrm{m}}$) $\in \mathbb{R}^{2d}$

\State \textbf{Step 2: MSE Loss Calculation}
\State $\mathcal{L}_{\text{Head1}} = \mathrm{MSE} \left( N_1(\text{Head1}) - \Delta T_{m} \right)$
\State $\mathcal{L}_{\text{Head2}} = \mathrm{MSE} \left( N_2(\text{Head2}) - \Delta T_{m} \right)$
\State where $N_1$ and $N_2$ are linear layers connected after Head1 and Head2 respectively. MSE is Mean Squared Error loss.

\State \textbf{Step 3: Ensemble}
\State $\hat{y}_{\text{ensemble}} = \frac{1}{2}(N_1(\text{Head1}) + N_2(\text{Head2}))$
\State $\mathcal{L}_{\text{ensemble}} = \frac{1}{2} \mathrm{MSE}(\hat{y}_{\text{ensemble}} - \Delta T_{m})$

\end{algorithmic}  
\end{algorithm}

We also select other protein language models as feature extractors for comparison. As shown in Figure \ref{fig:openfold_esm}, ESM2-650M \cite{lin2022language} and SaProt-650M \cite{su2023saprot} have architectures similar to ESM3-1.4B but accept only sequence inputs. Notably, SaProt relies on Foldseek \cite{van2022foldseek} to obtain structure-aware sequences as input, so we need an additional process when building \textit{Saprot-DTm}. OpenFold \cite{ahdritz2024openfold} is another feature extractor we employ. We extract sequence features $P_w$ and $P_m$ from the Evoformer and Structure Module, and also the embeddings at the specific mutated positions, $a_w$ and $a_m$. These features are then pass through a linear layer.

We train the model using the Adam optimizer\cite{diederik2014adam} with a learning rate of \textbf{$1 *10^{-5}$} and the OneCycle scheduler for 10 epochs. Gradient clipping with a norm of 0.1 is applied to ensure stable training. For all protein language backbone, we did not freeze the transformer backbone and trained all model weights in an end-to-end manner. For openfold backbone, we freeze the backbone and only train the linear layer.
% \yanzeng{add validation part / other details (batch size and padding)?}

% \daiheng{explained in part 2.2, maybe extend later}

\subsection{Regression Head}
% \daiheng{yan zeng, help me draw a picture for regression head fusion for diff backbone here using notations above.}

\label{Re_head}
For ESM2 and ESM3 backbone, the feature extraction process mainly adopted the corresponding CLS embeddings for global information and mutated position embeddings for local information. We adopt two supervised fine-tuning ways to fusion the wild-type and mutated sequences:
\begin{itemize}
\item Outer product of $a_w$ and $a_m$.
    % \item Linear combination of mutation position embeddings.
    % \item Linear combination of CLS embeddings.
\item Linear combination of $cls_w$ and $cls_m$ concatenated with $a_w$ and $a_m$.
\end{itemize}
For the OpenFold backbone, since it does not provide CLS embeddings, we use the outputs of the entire sequence after the Evoformer and Structure Modules as global embeddings. We continue to use the embeddings at the mutated positions for local information. Next, We also adopt two supervised fine-tuning ways to fusion the wild-type and mutated sequences:
\begin{itemize}
    \item Outer product of $a_w$ and $a_m$.
 \item Linear combination of $P_w$ and $P_m$.
    % \item Linear combination of $cls_w$ and $cls_m$ concatenated with $a_w$ and $a_m$.
\end{itemize}
For the SaProt backbone, since it also does not provide CLS embeddings, we use the mean of the entire sequence embedding as global information and mutated position embeddings as local information. We also adopt two supervised fine-tuning ways to fusion the wild-type and mutated sequences:
\begin{itemize}
    \item Outer product of $a_w$ and $a_m$.
 \item Linear combination of $P_w$ and $P_m$.
    % \item Linear combination of $cls_w$ and $cls_m$ concatenated with $a_w$ and $a_m$.
\end{itemize}

% For SaProt backbone, since SaProt's architecture is based on the ESM model and also includes CLS tokens, we use the same fusion method as ESM above.

% \begin{figure*}[!t]
%     \centering
%     % First image
%     \label{fig:openfold_esm}
%     \includegraphics[width=0.8\textwidth]{AnonymousSubmission/LaTeX/openfold_esm.pdf}
    
%     \vspace{0.5em} % Optional: Adjusts the space between the two images
    
%     % Second image
%     \label{fig:ESM3}
%     \includegraphics[width=0.8\textwidth]{AnonymousSubmission/LaTeX/esm3.pdf}
    
%     \caption{
%         \textbf{Model architecture}. \textit{ESM3-DTm} efficiently predicts $\dTm$. We also present \textit{ESM2-DTm, Saprot-DTm} and \textit{Openfold-DTm} here. ''I4A'' means mutation from I to A at position 4.
%     }
    
% \end{figure*}

\section{Results}
We evaluate different methods on $\Delta T_{\mathrm{m}}$
 in Table \ref{tab:main_result}. We primarily used the Pearson correlation coefficient (PCC), root mean square error (RMSE), and mean absolute error (MAE) to assess model performance. The PCC measures the linear correlation between the predicted and true values, indicating the model's ability to rank mutations by their $\Delta T_m$ values. The RMSE quantifies how closely the predicted measurements align with the true measurements, while the MAE provides the average absolute difference between predicted and true values. Here we can see that \textit{ESM3-DTm} surpasses \textbf{Geostab} 6.4\% in PCC, 1.9\% in MAE, and 4.4\% in RMSE.

\begin{table}[htbp]
    \centering
    % \begin{adjustbox} % Adjust the width as needed
    \begin{tabular}{lccc}
        \toprule
        Method & r(\(\uparrow\)) & MAE(\(\downarrow\)) & RMSE(\(\downarrow\)) \\
        \midrule
        \multicolumn{4}{l}{\textbf{Struct-based Methods}} \\ 
        HoTMuSiC       & 0.33 & 5.70 & 8.41  \\
        AUTO-MUTE     & 0.29 & 5.79 & 8.50  \\
        GeoDTm-3D     & 0.47 & 5.31 & 8.03  \\
        \midrule
        \textbf{Seq-based Methods} \\
        GeoDTm-Seq    & 0.46 & 5.55 & 8.11  \\
        ESM3-DTm     & \textbf{0.50} & \textbf{5.21} & \textbf{7.68} \\
        \bottomrule
    \end{tabular}
    % \end{adjustbox}
    \caption{Comparison with existing models on the S571 dataset. Other results are quoted from \cite{xu2023improving}. }
    \label{tab:main_result}
\end{table}

Furthermore, we fine-tuned models using the ESM2, ESM3, SaProt, and OpenFold backbones with similar architectures to make a fair comparison of their representation abilities, as presented in Table \ref{tab:Backbone}. The results indicate that the multimodal ESM3 backbone achieves the highest performance, suggesting that incorporating structural information benefits the prediction. 

% However, the performance of \textit{SaProt-DTm} did not meet our expectations. We believe there are two possible reasons for this:

% \textbf{1. Different Folding Models:} SaProt was trained on datasets generated by AlphaFold, whereas we used datasets generated by ColabFold. The differences between these folding models may cause discrepancies in the input PDB structures, leading to suboptimal compatibility with SaProt's model.

% \textbf{2. Limitations of Foldseek:} The structure-aware tokens obtained from Foldseek may not accurately capture the changes caused by mutations because Foldseek is primarily designed for sequence alignment rather than for capturing structural variations, which transforms the structure into only 20 tokens. For mutation prediction, this level of granularity may be too coarse, resulting in poor alignment and, consequently, causing the structure to have a negative impact. 

\begin{table}[!htbp]
    \centering
    \setlength{\tabcolsep}{0.4em} % Adjusts the horizontal padding between columns
    
    \begin{tabular}{lccc}
        \toprule
        Backbone & r(\(\uparrow\)) & MAE(\(\downarrow\)) & RMSE(\(\downarrow\)) \\
        \midrule
        \textbf{Seq-based Methods} \\
        ESM2-DTm & 0.48 & 5.31 & 7.85 \\
        ESM3-DTm (seq only) & 0.49 & 5.28 & 7.77 \\
        \midrule
        \textbf{Multimodal Methods} \\
        ESM3-DTm & \textbf{0.50} & \textbf{5.21} & \textbf{7.68} \\
        SaProt-DTm & 0.46 & 5.51 & 8.00 \\
        OpenFold-DTm & 0.35 & 5.31 & 8.03 \\
        \bottomrule
    \end{tabular}
 
    \caption{Comparison of different backbones on the S571 dataset.}
    \label{tab:Backbone}
\end{table}

% Here we refrain from excessive fine-tuning and aim to fairly compare the ability of various backbones' representations to evaluate DTm tasks under the same or similar settings.

% \daiheng{the explanation for this part is that if OpenFold results are good, it demonstrates the importance of structural information. A purely sequence-based model is not sufficient (as referenced by "mutate everything"). However, models like SaProt, which use Foldseek to create a language model for quantization, lose too much information through discretization, so their results are not good either.}

\section{Ablation Study}
Here we explore the performance of different regression heads. Using the ESM2-650M model as the backbone, we conducted all experiments under consistent settings, including those previously established. The results are presented in Table \ref{tab:Heads_compare_ESM2}. Based on these findings, we selected the best of three to form our final model. Additionally, we investigated the effects of fine-tuning versus freezing the ESM2-650M backbone during prediction, as shown in Table \ref{tab:Finetuning_Strategies}. Our results indicate that fine-tuning the backbone significantly improves prediction performance.
\begin{table}[htbp]
    \centering
    \setlength{\tabcolsep}{0.7em} % Adjusts the horizontal padding between columns
        \begin{tabular}{lcc}
            \toprule
            Regression Heads & r(\(\uparrow\)) & MAE(\(\downarrow\)) \\
            \midrule
            MUT Token concatenation & 0.21 & 6.34 \\
            MUT Token outer product & 0.41 & 5.69 \\
            MUT Token linear combination & 0.40 & 5.50 \\
            CLS Token linear combination & 0.33 & 6.01 \\ % Verify if this row is intentional
            % Third and fourth strategy concatenation & \textbf{0.48} & \textbf{5.31} \\
            \bottomrule
        \end{tabular}
    \caption{Comparison of different regression heads on the ESM2 Backbone.}
    \label{tab:Heads_compare_ESM2}
\end{table}

\begin{table}[htbp]
    \centering
      \setlength{\tabcolsep}{0.75em}
    % Place label immediately after caption

        % Adjust horizontal padding between columns
        \begin{tabular}{lccc}
            \toprule
            Fine-tuning Option & r & MAE & RMSE \\
            \midrule
            Fully fine-tune & \textbf{0.48} & \textbf{5.31 } & \textbf{7.85} \\
            Freezing backbone  & 0.46 & 5.50 & 7.89 \\
            \bottomrule
        \end{tabular}
    \caption{Comparison of finetuning strategies on ESM2 backbone.}
     \label{tab:Finetuning_Strategies}
\end{table}

We found that in the comparison of protein language model backbones, the results of SaProt are slightly lower than ESM2 and ESM3. Since SaProt does not have a CLS token, we also use the same combination of $a_w$ and $a_m$ for ESM2 when comparing it with SaProt. The result is shown in Table \ref{tab:avg_compare}. One possible explanation for this result is that SaProt was trained on datasets generated by AlphaFold, whereas we used datasets generated by ColabFold. The differences between these folding models may cause discrepancies in the input PDB structures, leading to suboptimal compatibility with SaProt's model. Another possible reason is that the structure-aware tokens obtained from Foldseek may not accurately capture the changes caused by mutations because Foldseek is primarily designed for sequence alignment rather than for capturing structural variations, which transforms the structure into only 20 tokens. For mutation prediction, this level of granularity may be too coarse, resulting in poor alignment and, consequently, causing the structure to have a negative impact.

\begin{table}[!htbp]
    \centering
      \setlength{\tabcolsep}{0.75em}
    % Place label immediately after caption

        % Adjust horizontal padding between columns
        \begin{tabular}{lccc}
            \toprule
            Backbone & r & MAE\\
            \midrule
            ESM2 &  0.30& 5.84 \\
            SaProt  &0.15  & 6.25  \\
            \bottomrule
        \end{tabular}

    \caption{Linear combination of avg pooling.}
     \label{tab:avg_compare}
\end{table}

\section{Conclusion}
In this work, we propose to utilize multimodal protein language model backbone to make effective prediction on melting temperature.  Our findings demonstrate that the multimodal model \textit{ESM3-DTm} outperforms other single-modality models. By effectively incorporating both sequence and structural data in fully fine-tuning strategy, we can achieve more accurate predictions.

\section{Acknowledgments}
We would like to acknowledge the support from Beijing Municipal Science \& Technology Commission, Administrative Commision of Zhongguancun Science Park (Z221100003522019).

\bibliography{aaai25}

\end{document}